\documentclass[graybox]{svmult}

\usepackage{type1cm}        
\usepackage{makeidx}         
\usepackage{graphicx}        
\usepackage{multicol}        
\usepackage[bottom]{footmisc}

\usepackage{newtxtext}       %
\usepackage{newtxmath}       
\usepackage{amsmath}
\usepackage{multirow}
\usepackage{natbib}
\usepackage{url}
\newcommand{\modelFive}[0]{FreezeB$5$}
\newcommand{\modelFour}[0]{FreezeB$4$}
\newcommand{\modelThree}[0]{FreezeB$3$}
\newcommand{\modelTwo}[0]{FreezeB$2$}
\newcommand{\modelOne}[0]{FreezeB$1$}
\newcommand{\modelZero}[0]{FreezeB$0$}

\makeindex             

\begin{document}
\title*{Do Deep Neural Networks Forget Facial Action Units? - Exploring the Effects of Transfer Learning in Health Related Facial Expression Recognition}
\titlerunning{Do Deep Neural Networks Forget Facial Action Units?}

\author{
    Pooja Prajod and  
    Dominik Schiller and
    Tobias Huber and 
    Elisabeth André
}

\institute{
Pooja Prajod \at Human Centered Multimedia, Augsburg University, Augsburg, Germany\\ \email{pooja.prajod@uni-a.de}
\and Dominik Schiller \at Human Centered Multimedia, Augsburg University, Augsburg, Germany\\ \email{dominik.schiller@uni-a.de}
\and Tobias Huber \at Human Centered Multimedia, Augsburg University, Augsburg, Germany \\ \email{tobias.huber@uni-a.de}
\and Elisabeth André \at Human Centered Multimedia, Augsburg University, Augsburg, Germany\\ \email{andre@informatik.uni-augsburg.de}
}

\maketitle

\abstract{
In this paper, we present a process to investigate the effects of transfer learning for automatic facial expression recognition from emotions to pain. 
To this end, we first train a VGG16 convolutional neural network to automatically discern between eight categorical emotions. 
We then fine-tune successively larger parts of this network to learn suitable representations for the task of automatic pain recognition.
Subsequently, we apply those fine-tuned representations again to the original task of emotion recognition to further investigate the differences in performance between the models.
In the second step, we use Layer-wise Relevance Propagation to analyze predictions of the model that have been predicted correctly previously but are now wrongly classified.
Based on this analysis, we rely on the visual inspection of a human observer to generate hypotheses about what has been forgotten by the model. 
Finally, we test those hypotheses quantitatively utilizing concept embedding analysis methods.  
Our results show that the network, which was fully fine-tuned for pain recognition, indeed payed less attention to two action units that are relevant for expression recognition but not for pain recognition.}

\section{Introduction}

Facial expressions play a crucial part in the nonverbal communication of every social interaction.
They can convey a persons inner state or intentions without the need of verbalization, which is especially useful for people that are not able to express themselves using speech (e. g. because of illness or accidents). 
This makes the assessment of a patient's facial expressions an important aspect to consider in many healthcare applications and also a valuable skill to learn for staff in hospitals and nursing homes. 
In everyday nursing home life for example, pain recognition is an important routine task for nursing staff in order to enable optimal pain management through individually adapted therapy.
Additionally to the direct assessment of pain, facial expressions can provide valuable insights about associated emotional states like panic or confusion.

However, it is practically impossible for medical staff to continuously monitor patients manually. 
As a consequence, there has been an growing interest in the research community to develop methods that automatically recognize if a person is in pain or in an associated emotional state.
The SenseEmotion research project \cite{velana2016senseemotion} for example investigated the optimization of pain treatment through automatic detection of physical pain and the reduction of mental pain through affect management.
In a similar way the KRISTINA project \cite{wanner2017kristina} aimed at automatically recognizing  the emotions of elderly people in nursing homes.

To this end, modern state of the art approaches often rely on deep learning techniques that are able to learn a suitable representation for a given task from the raw data input \cite{luqin19asurvey, li2020deep}. 
While those methods often achieve superior classification performance over traditional handcrafted features, they require large amounts of annotated training data to do so.  
This provides a challenging environment for the classification of sensitive tasks, like automatic pain recognition, where the availability of data is limited due to sparsity of patient contact, privacy concerns \cite{cowie2017electronic}, and due to the adherence of strict ethical guidelines \cite{charlton1995ethical}.
A frequently used approach to facilitate the learning process in situations with scarce data comes in the form of transfer learning.
Here the principal idea is to transfer the knowledge that a model has learned for a specific task A over to an adjacent task B. 
While such a transfer can contribute to an increase in performance for task B it is a common phenomenon that the model forgets crucial information, which it has previously learned for task A  \cite{yosinski2014transferable}. 
In practical applications however, it is often desirable to learn representations of the input data that can be used for multiple tasks at the same time.

In this paper, we propose a novel, multi-step approach to investigate what neural networks forget during the process of transfer learning from emotions to pain. 
In the first step we evaluate the difference in emotion recognition performance before and after the transfer learning process. 
In the second step, we use explainable AI methods to analyze predictions of the model, that have been predicted correctly previously but are now wrongly classified.
Based on this analysis we rely on the visual inspection of a human observer to generate hypotheses about what has been forgotten by the model. 
Finally, we test those hypotheses quantitatively utilizing concept embedding analysis methods.  

\section{Related Work}

\subsection{Transfer Learning in Pain Recognition}

Transfer learning is a widely used technique to circumvent the problem of small sized datasets.  
In this section, we discuss some of the works that propose facial expression based pain recognition using transfer learning.

One of the earlier works on pain recognition using transfer learning was by \cite{florea2014learning}. 
They used hand-crafted Histograms of Topographical (HoT) features to learn a suitable data representation for emotion recognition. 
They transferred the learned emotion representation to fit a support vector regressor for estimating pain intensity, thereby achieving an overall improvement in robustness and generalization capabilities of the system. 

However, \cite{zamzmi2018neonatal} and \cite{egede2017fusing} demonstrated that deep neural networks outperform traditional hand-crafted features in automatic pain recognition.
They extracted deep learned features from pre-trained convolutional neural networks (CNN) and fused them with hand-crafted features. 
The fused feature representation was used to train classifiers for recognizing pain. 
Both works obtained a significant improvement over hand-crafted features.

While Zamzmi et al and Egede et al transferred learned representations from CNNs by using the network as feature extractors, \cite{wang2017regularizing} adopted a different approach. 
In their experiments they used a fine-tuning technique to leverage the learned representations from a pre-trained CNN model for pain intensity estimation. 
Wang et al demonstrated that fine-tuning a pre-trained CNN from a data-rich domain (e.g. face verification) can mitigate the problem of fully training a deep learning network on small datasets (e.g. pain recognition).

Most of the existing pain datasets are composed of consecutive video frames and contain temporal information that is often not utilized.
This led to the idea of using recurrent neural networks.
\cite{rodriguez2017deep} fine-tuned a pre-trained CNN on a pain dataset and used it as a feature extractor.
They then used those features as input, to train a Long Short Term Memory network (LSTM) for a binary pain recognition task. 
Adding the the temporal information, they could achieve superior performance over just using the frame wise CNN representations.
A study by \cite{haque2018deep} however showed, that the exploitation of the temporal information does not always improve results.
In their experiments they fine-tuned three CNNs on pain intensity, using RGB, depth information, and thermal video frames as input.
They concluded that, given the limited number of training sequences, the frame-wise CNN features were not sufficiently discriminative to obtain good LSTM generalizations.

Fine-tuning based transfer learning often changes the internal representations which were learned for the original task. 
These changes may result in forgetting of crucial information that reduces the performance of the model on the original task. 
The aforementioned studies focused on fine-tuning models to obtain the best result on the target task. 
However, they did not delve into the consequences of the fine-tuning process on the performance of the original task.
\cite{kemker2018measuring} investigated forgetting within a single task.
However, as far as we know there is no work on quantifying forgetting between distinct sequentially learned tasks.

\subsection{Explainable Artificial Intelligence}

With the increasing popularity of fine-tuning techniques in high-risk domains like health-care, it is important to investigate what is learned and what is forgotten. 
A common way of analyzing the predictions of neural networks is the creation of so called \emph{saliency maps} that highlight how important each input was for the prediction.
For pain recognition, \cite{weitz2019deep} applied and compared Layer-wise Relevance Propagation (LRP) (\cite{montavon2019lrp_overview}) and Local Interpretable Model-agnostic
Explanations (LIME) (\cite{ribeiro2016lime}).
Weitz et al conclude that the salience maps generated by those approaches already provide some insights into the reasoning of the network but are not expressive enough on their own since they are often ambiguous and hard to interpret for end-users.

To combat this problem, recent work on \emph{concept embedding analysis} investigates which human-comprehensible concepts were learned by a given network.
\cite{bau2017networkDissection} showed that there are often semantic concepts embedded in single neurons of the latent space of a neural network.
For instance, \cite{khorrami2015deep} showed that certain neurons inside the last convolutional layer of a network, which was trained for facial expression analysis, learned to recognize specific Action Units (AUs) - visible indicators of the operation of individual facial muscles.
To extend this method to account for concepts that might span over several neurons in the latent space, \cite{kim2018interpretability} proposed to train a binary linear classifier that takes the output of a intermediate layer of the network as input and learns to recognize a given concept.
If the linear classifier is able to identify the concept then it is likely that the network learned this concept.
They tested their approach on several image classification models and a model for predicting diabetic retinopathy.
Similarly, \cite{schwalbe2020concept} used convolutional networks instead of a linear classifier to search for segmented concepts in the intermediate layers. 
A common challenge for the aforementioned approaches is that the potential concepts have to be externally identified by human experts.

In this work, we utilize LRP saliency maps to facilitate the identification of potential concepts.
Then we follow the approach by \cite{kim2018interpretability} to verify whether the network learned those concepts.
As far as we know, concept embedding analysis has not yet been explored for facial expression recognition tasks. 
However, healthcare related applications like pain recognition would benefit from the additional understanding of the models inner working.

\section{Pain Training}

Usually, pain datasets are small and have very few pain samples \cite{wang2017regularizing}. 
In such cases, a transfer learning approach is often adopted. 
It involves re-using some parameters from a pre-trained model and training the remaining parameters on a small target dataset. 
Ideally, the pre-trained model is trained on a large dataset from a domain similar to the target domain. 
Emotion recognition from facial expression is a well studied task with large existing datasets like AffectNet \cite{mollahosseini2017affectnet}. 
Since facial expressions, in particular AUs, are used in both pain and emotion recognition \cite{simon2008recognition}, we choose facial emotion recognition as our source task.

\subsection{Datasets}
\label{sec:datasets}

We use \textbf{AffectNet} \cite{mollahosseini2017affectnet} for the facial emotion recognition task. 
The AffectNet dataset contains $420299$ manually annotated face images belonging to 11 classes - Neutral, Happy, Sad, Surprise, Fear, Disgust, Anger, Contempt, None, Uncertain, Non-Face. 
The images were collected through search queries containing emotional keywords. 
This dataset is imbalanced with very high number of samples ($> 75000$) in categories like 'Happy' and 'Neutral', and lower samples (around $4000$) in 'Disgust' and 'Contempt'. 
We excluded images belonging to ‘None’, ‘Uncertain’ and ‘Non-face’ as they do not have an emotion label. 
From the remaining, we use  $273269$ images for training and $14382$ for validation. 
As suggested by the authors, we use the validation images from the original dataset as test set. 
Hence, we have $4000$ images ($500$ per class $\times$ $8$ classes) for testing. 

For pain recognition, we use the face images from the \textbf{UNBC-McMaster} shoulder pain expression database \cite{lucey2011painful}. 
The images were generated from video recordings of $25$ participants who suffered from shoulder pain. 
Each image has a Prkachin and Solomon Pain Intensity (PSPI) score, where $0$ indicates no-pain and scores $> 0$ indicate different levels of pain.
The dataset has $48398$ ($40029$ no-pain $+\ 8369$ pain) images. 
We use all images from $4$ participants (allowed to publish) as test set.
We want to simulate a transfer learning use-case with very few samples available in the target domain. 
Hence, $1000$ images ($500$ pain $+\ 500$ no-pain) are selected from the remaining $21$ participants for the training ($900$ images) and validation ($100$ images) phase of transfer learning.
To avoid very similar video frames, the training and validation images are chosen randomly but satisfy the conditions that (i) there is at least 1 image of pain and no-pain from each of the $21$ participants, (ii) the set has no consecutive images.

\subsection{Transfer Learning}
\label{sec:emotrain}
As noted by \cite{yosinski2014transferable}, there are two methods for transfer learning in deep neural networks - freezing and fine-tuning. 
In freezing method, weights of first few layers are copied from a pre-trained model and these layers are marked as frozen or not trainable. Only the remaining unfrozen layers are trained on target dataset. 
Similar to freezing, fine-tuning also involves copying weights from pre-trained model. The difference is that no layer is marked frozen, i.e. all layers are further trained on the target dataset. 

We adopt a hybrid method which involves: (i) training a base model on the source dataset, (ii) copying all the weights from the base model to the target model, (iii) freezing some layers and fine-tuning the remaining layers on the target dataset. 

We use the VGG16 architecture with 5 convolution blocks for both emotion and pain recognition tasks. All images from both datasets are scaled to default VGG16 input dimensions (224 $\times$ 224). In both models, we use SGD optimizer (learning rate $= 0.01$), focal loss and data augmentation. The focal loss function \cite{lin2017focal} is given by: 
\begin{equation}
    focal\_loss = (1 - p_t)^\gamma \times cross\_entropy\_loss
    \label{eq:fl}
\end{equation}
$p_t$ is the predicted probability of a sample belonging to its true class ($t$) and $\gamma$ is a hyperparameter. 

For the first step, we train an emotion recognition model using images from the AffectNet dataset (see section \ref{sec:datasets}). We use a VGG16 model pre-trained on ImageNet \cite{deng2009imagenet} as the starting point for the emotion recognition task. All the layers are marked trainable. 
The model is eventually connected to a dense layer with softmax activation to predict the probability of an image belonging to each of the 8 emotion classes - Neutral, Happy, Sad, Surprise, Fear, Disgust, Anger and Contempt. During training, we use horizontal flip to augment the data. We set the hyperparameter in focal loss function (equation \ref{eq:fl}) as $\gamma = 5$. Since our training and validation sets are heavily imbalanced, we use a weighted focal loss function. We use the weighting scheme proposed by \cite{cui2019class}, given by:
\begin{equation}
    weighted\_loss = \frac{1-\beta}{1-\beta^{samples\_per\_cls}} \times focal\_loss
\end{equation}
We set the hyperparameter $\beta = 0.99998$.
\begin{figure}[t]
\sidecaption[t]
  \centering
  \includegraphics[width=0.649\linewidth]{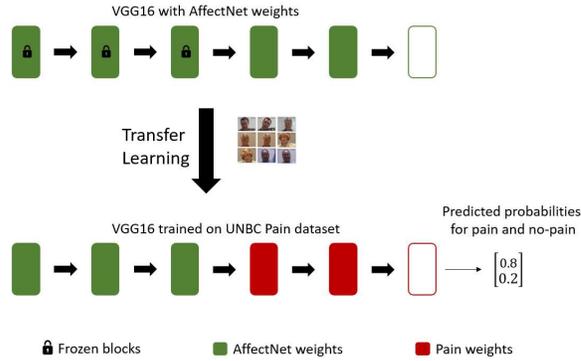}
  \caption{Illustration of the process we followed for transfer learning from emotion to pain. This images shows the process of generating the model \modelThree{} by freezing the first 3 blocks of the emotion model.}
  \label{fig:paintrain}
\end{figure}

We use the emotion recognition model trained on AffectNet as the source model for transfer learning pain (figure \ref{fig:paintrain}). We copy the weights of all convolution blocks and freeze the first few blocks, making them unavailable for pain training. We vary the number of frozen blocks from 0 (all blocks are available for pain training) to 5 (none of the convolution blocks are available for pain training, only the output layers are trained). This yields six pain recognition models. These models are trained on a variant of the UNBC shoulder pain dataset (see section \ref{sec:datasets}) to determine if a face image shows pain (PSPI pain score $> 0$) or not (PSPI pain score $= 0$). 
For this task, we use Keras data augmentation options: rotation ($[-25^o, 25^o]$), height shift ($[-10\%, 10\%]$), width shift ($[-10\%, 10\%]$), shear ($[-10\%, 10\%]$), zoom ($[-10\%, 10\%]$) and horizontal flip.
Since our training and validation sets are balanced, we use the unweighted focal loss (equation \ref{eq:fl}) with $\gamma = 2$.

\section{Measuring Forgetting}

Catastrophic forgetting occurs when transfer learning causes large changes in weights of the existing model and thus altering previous feature representations (\cite{kemker2018measuring}). 
In our case, when more unfrozen layers are available for training pain, the model becomes increasingly tailored for pain recognition. 
We say our model forgot the recognition of an emotion if the recall of the emotion reduces after transfer learning. 

\subsection{Re-train for Measuring Forgetting}
\label{sec:retrain}
Since we vary the number of frozen blocks in pain training from 0 to 5, we have 6 different pain recognition models.
The models are labelled 'FreezeB$\langle i \rangle$', where $i= 0$ to $5$. 
'FreezeB$\langle i \rangle$' is generated by freezing the first $\langle i \rangle$ convolution blocks with initial emotion recognition weights and training the remaining blocks for pain recognition. 
'\modelFive{}' corresponds to the model where all convolution blocks are frozen and only the output layers are trained for pain recognition. 
On the other hand, '\modelZero{}' corresponds to the pain model generated by not freezing any blocks, i.e., all the layers were trained on pain dataset.
Evaluation measures, like recalls, help in determining the best pain recognition model. 
We also want to evaluate these models in terms of forgetting and study the trade-off between pain recognition and emotion recognition. 
In other words, we want to determine how these models perform on emotion recognition after feature representations are adapted for pain recognition.

\begin{figure}[t]
\sidecaption[t]
  \centering
  \includegraphics[width=0.649\linewidth]{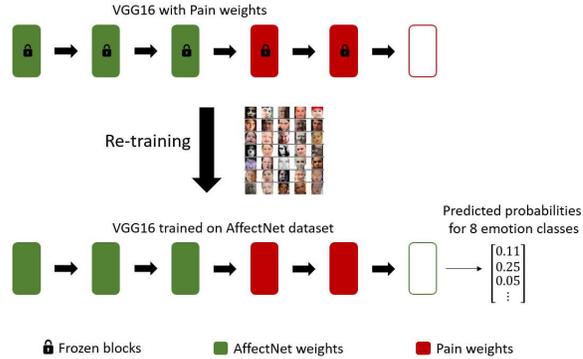}
  \caption{Illustration of the process we followed for re-training the various pain models. All convolution blocks are frozen and only the output layers are trained to predict emotions instead of pain.}
  \label{fig:retrain}
\end{figure}

To measure forgetting we employ a re-training methodology (figure \ref{fig:retrain}), i.e., we evaluate the capability of the pain models to recognize emotions instead of pain. 
First, we freeze all convolution blocks of a pain model so that the learned feature representations are intact. 
Next, we add output layers to get prediction probabilities for 8 emotion classes. 
Finally, we train the model on the AffectNet dataset with the same optimizer, data augmentation, loss function and hyperparameters as before (see section \ref{sec:emotrain}). 
This process is repeated for all 6 pain models and the resulting models are compared based on the their recalls for each emotion. 
We run McNemar's test to determine if there are any significant difference in recalls (\cite{dietterich1998approximate}).

\subsection{Visual Analysis}
\label{sec:xai}

\begin{figure}[t]
    \sidecaption
    \begin{minipage}[t]{0.64\linewidth}
        \begin{minipage}[t]{0.32\linewidth}
        \centering
            \includegraphics[width=\linewidth]{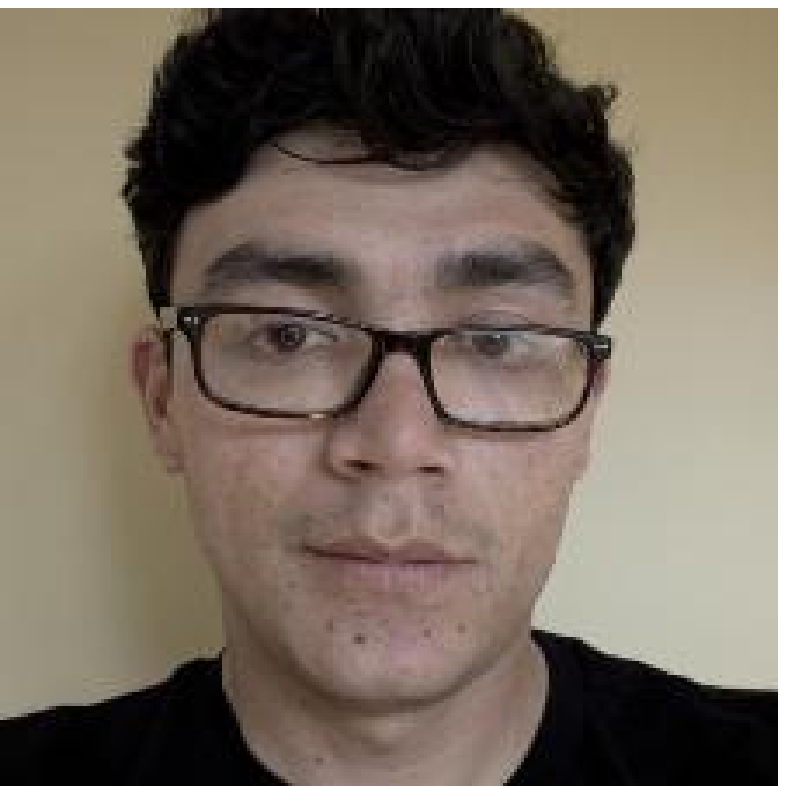}
            \parbox[t][0.5\baselineskip]{\linewidth}{
            \centering
            Input Image}
        \end{minipage}
        \begin{minipage}[t]{0.32\linewidth}
        \centering
            \includegraphics[width=\linewidth]{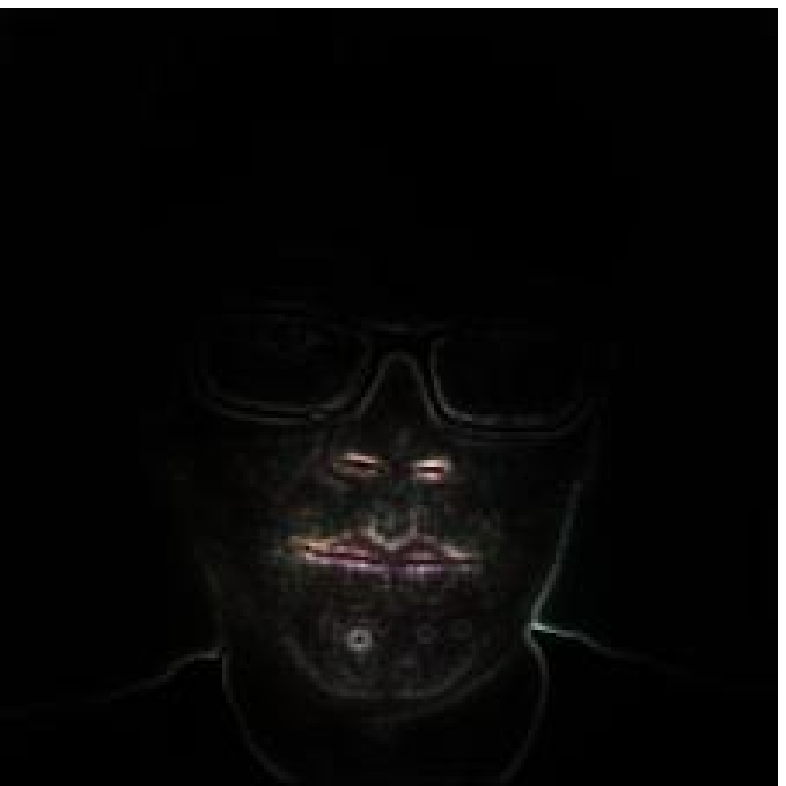}
            \parbox[t][0.5\baselineskip]{\linewidth}{\centering \modelFive{}}
        \end{minipage}
        \begin{minipage}[t]{0.32\linewidth}
        \centering
            \includegraphics[width=\linewidth]{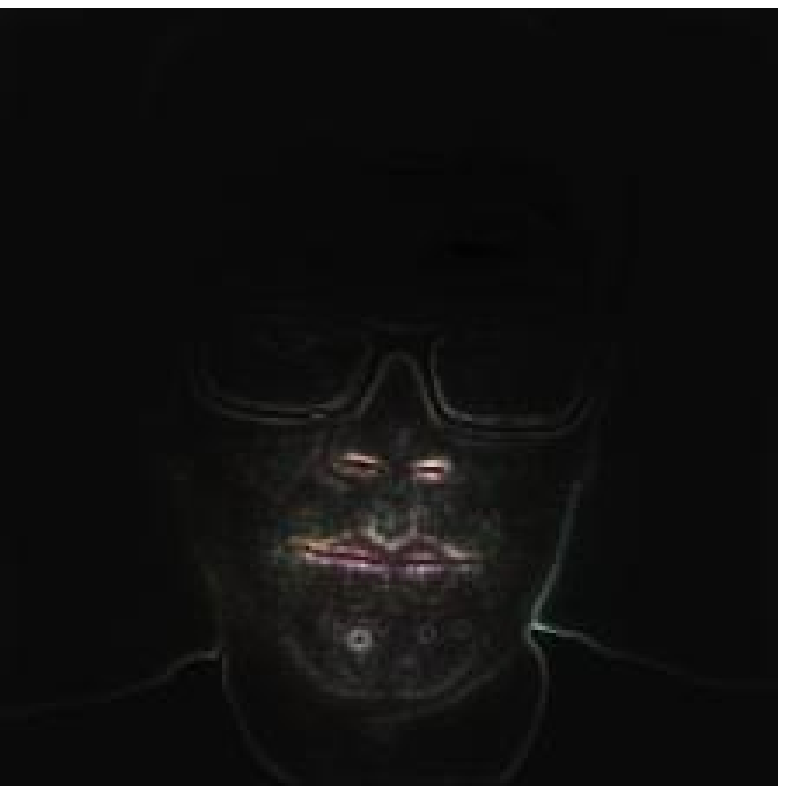}
            \parbox[t][0.5\baselineskip]{\linewidth}{\centering \modelZero{}}
        \end{minipage}
    \end{minipage}
    \caption{Example for an image that \modelFive{} predicted correctly as \emph{contempt} but \modelZero{} did not. 
    The middle and right pictures display saliency maps that highlight which regions influenced each model to predict \emph{contempt}.}
    \label{fig:example_saliency}
\end{figure}

While the quantitative evaluation of the emotion recognition models allows a performance comparison between them, we do not know which specific embedded concepts might have been learned or forgotten to cause any difference in performance.
To analyze the difference between \modelFive{}, which uses the same features as the original emotion model, and one of the fine-tuned models (say FreezeB$\langle k \rangle$, with $ k \in \{0, 1, 2, 3, 4\}$), we look at the emotion images that got predicted correctly by \modelFive{} but wrongly by FreezeB$\langle k \rangle$.
For both of the models, we create a saliency map that highlights the areas of the input image which have the most influence on the model to correctly classify a given sample. 
The saliency maps are created using LRP with the $z$-rule for fully connected layers and the $z^+$-rule for convolution layers.
This is recommended by \cite{montavon2019lrp_overview} and was shown to be less independent of the models learned parameters than other LRP methods by \cite{sixt2019explanations}.
Additionally, the saliency maps are normalized between $0$ and $1$, such that a value of $0.5$, for example, indicated that this pixel accounted for half of the confidence in the prediction. 
Since both models were fine-tuned based on the same emotion recognition weights, the saliency maps for the same input image and the same class often appear indistinguishable to the human eye (see figure \ref{fig:example_saliency}).

Unlike humans however, neural networks can be sensitive to very small differences.
To make those differences visible, we generate new saliency maps by subtracting the raw saliency maps from each other and normalizing these differences between $0$ and $1$ for visibility. 
We use them to identify concepts which \modelFive{} payed more attention to than FreezeB$\langle k \rangle$.
The underlying intuition is that those concepts helped \modelFive{} to correctly predict the emotion but were not used enough by FreezeB$\langle k \rangle$ (i.e. were forgotten by FreezeB$\langle k \rangle$).
To translate relevant image areas into semantically meaningful concepts, the interpretation needs to be done visually by a human. 
Hereby, we especially focus action units as candidates for concepts, since they are known to be good indicators for emotion as well as pain and have already been shown to be utilized by neural networks (\cite{khorrami2015deep}).

\subsection{Concept Embedding Detection}
\label{sec:concept_embedding}

In this section, we describe how we verify the hypothesis that the concept candidates identified in section \ref{sec:xai} were actually forgotten after fine-tuning our model for pain recognition. 
To this end, we investigate whether these concept candidates are embedded in the latent space of the models by applying concept embedding detection.
One way to detect a specific concept, is training a binary linear classifier on the output of an intermediate layer of the network to recognize this concept (\cite{kim2018interpretability}).
If the linear classifier is able to detect the concept then it is likely that the network learned this concept.

We focus on AUs as potential concepts but the AffectNet dataset does not contain labeled AUs.
According to \cite{kim2018interpretability}, concept detection need not be done on the dataset on which the model was trained. 
Hence, we use the CK+ dataset (\cite{lucey2010ckplus}), which has been specifically developed to serve as a comprehensive test-bed for comparative studies of facial expression analysis.
This dataset consists of video recordings of acted emotions, where the frame displaying the peak of the emotion has been manually annotated with respect to action units, which makes it well suited for our case. 
By computing the output of the last convolution block of \modelFive{} and FreezeB$\langle k \rangle$ ($k \in \{0, 1, 2, 3, 4\}$), we obtain two feature-sets that represent the latent space of the respective models.
We then train a linear Support Vector Machine (SVM) to detect the concept candidate on each of those two feature-sets using 2-fold cross validation.
For each feature-set, we compute the average f1-score of the two folds. 
We repeat this training process for 500 iterations using different random seeds for weight initialisation and fold image selection. 
Finally, we run a paired t-test between the 500 averaged F1-scores of each feature-set.
The result of this test shows whether there is a significant difference between the performance of SVMs trained on the \modelFive{} features and the SVMs trained on the FreezeB$\langle k \rangle$ features. 
\cite{dietterich1998approximate} suggests this comparison metric for $5$ iterations as $5 \times 2$ cross validation paired t-test and we extend it to 500 iteration as suggested by \cite{kim2018interpretability}.

\section{Results}

Our pain training procedure yields 6 different models. 
First, we compare the performance of these models in recognizing pain. 
Next, we evaluate these models in terms of forgetting emotion recognition. 
These metrics help us to assess the suitability of the learned representations for emotion as well as for pain recognition..
Additionally, we utilize XAI to analyze which embedded concepts (e.g. action units) are forgotten during transfer learning from emotions to pain.

\begin{table*}[t]
\caption{Class-wise precision, recall and F1-score along with the respective macro averages for each of the pain models}
\begin{center}
\begin{tabular}{ |c|c|c|c|c|c|c|c|c|c|c| } 
\hline
\multirow{2}{6em}{Pain models}  & \multicolumn{3}{|c|}{Precision} & \multicolumn{3}{|c|}{Recall} & \multicolumn{3}{|c|}{F1-score} & \multirow{2}{4em}{Accuracy}\\
\cline{2-10}
& No pain & Pain & Avg. & No pain & Pain & Avg. & No pain & Pain & Avg. & \\ 
\hline
\modelFive{} & 0.92 & 0.87 & 0.90  & \textbf{0.99} & 0.54 & 0.76  & 0.95 & 0.66 & 0.81 & 0.92 \\ 
\modelFour{} & 0.93 & \textbf{0.90} & \textbf{0.92}  & \textbf{0.99} & 0.56 & 0.78  & \textbf{0.96} & 0.69 & 0.83 & 0.92 \\ 
\modelThree{} & \textbf{0.95} & 0.85 & 0.90  & 0.98 & 0.69 & 0.83  & \textbf{0.96} & \textbf{0.76} & \textbf{0.86} & \textbf{0.93} \\ 
\modelTwo{} & \textbf{0.95} & 0.75 & 0.85  & 0.96 & \textbf{0.71} & \textbf{0.84}  & 0.95 & 0.73 & 0.84 & 0.92 \\ 
\modelOne{} & \textbf{0.95} & 0.72 & 0.83  & 0.95 & \textbf{0.71} & 0.83  & 0.95 & 0.72 & 0.83 & 0.91 \\ 
\modelZero{} & \textbf{0.95} & 0.72 & 0.83  & 0.95 & \textbf{0.71} & 0.83  & 0.95 & 0.71 & 0.83 & 0.91 \\ 
\hline
\end{tabular}
\end{center}
  \label{tab:painperformances}
\end{table*}

Table \ref{tab:painperformances} lists the performance of the 6 pain recognition models in terms of precision, recall, f1-score and accuracy. 
Accounting for the imbalance in our test set for pain recognition (see section \ref{sec:datasets}), we use macro average of the performance metrics (compute the metric for each class and average them) as it treats every class equally. 
As expected, models with higher number of convolution blocks available for pain training (FreezeB$\langle i \rangle,\ i = 0,1,2$) have higher pain recall. 
The pain recalls saturate beyond a point, i.e., unfreezing blocks for pain training beyond a point (in our case \modelTwo{}) does not yield better pain recall. 
\begin{figure}
  \centering
  \includegraphics[width=\linewidth]{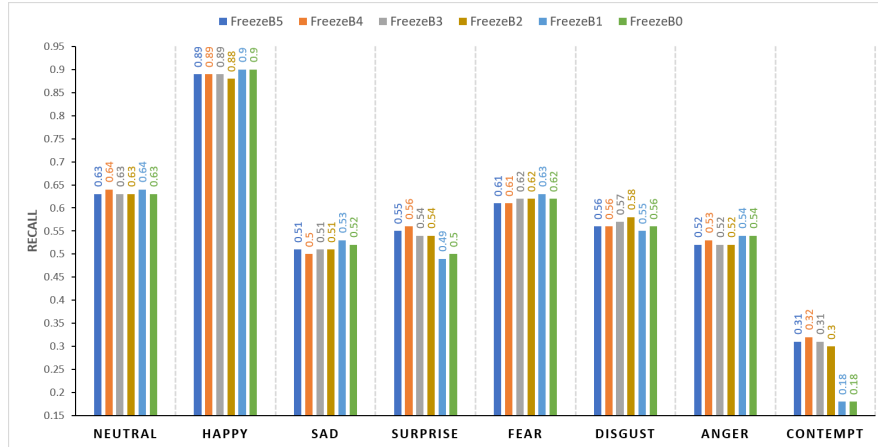}
  \caption{Class-wise recalls of the six pain models obtained by re-training their output layers for emotion recognition.}
  \label{fig:retrainemorecall}
\end{figure}
After generating different pain models, we freeze all convolution blocks of the model and re-train only the output layers for emotion recognition.
This preserves the learned feature representations after pain training and helps measure it's impact on emotion recognition. 
Figure \ref{fig:retrainemorecall} shows the recalls of the 8 emotions when using the learned representation of the 6 pain recognition models. The recalls of surprise and contempt reduce over the blocks and is notably lower for \modelOne{} and \modelZero{}. 
For demonstration of our approach, we choose \modelZero{} for further analysis.
Using McNemar's test between \modelFive{} and \modelZero{}, we found that the difference in recall is significant for both surprise (p-value: $1.56\mathrm{e}{-4}$) and contempt(p-value: $8.19\mathrm{e}{-18}$).

\begin{figure}
    \sidecaption[t]
    \centering
    \includegraphics[width=0.649\linewidth]{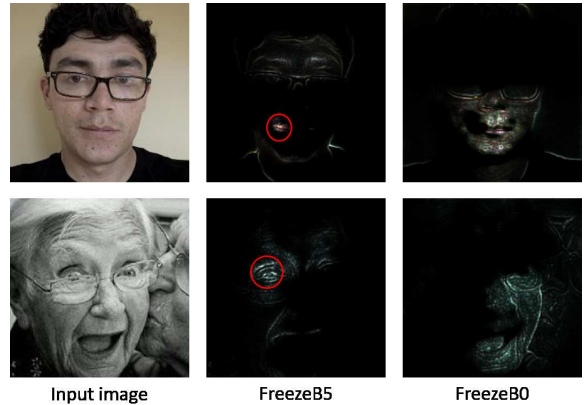}
    \caption{Visualization of the difference between the saliency maps of \modelFive{} and \modelZero{}  for contempt (top) and surprise (bottom) images.
    The red circles highlight the areas that correlate with the action units which \modelFive{} payed more attention to.
    }
    \label{fig:saleincy_difference_contempt}
\end{figure}
To investigate what concepts might have been forgotten, we generated saliency maps for images that were correctly classified as contempt or surprise by \modelFive{} but wrongly classified by \modelZero{}, as described in section \ref{sec:xai}.
An example image for each of the two emotions and the difference in their saliency maps can be seen in figure \ref{fig:saleincy_difference_contempt}.
A visual comparison between saliency maps for all images in the test set from the contempt class (see section \ref{sec:xai}) indicates that \modelFive{} is paying more attention to dimples in the face than \modelZero{} (see figure \ref{fig:saleincy_difference_contempt} first row). 
Since dimples corresponds to AU14, this leads us to the hypothesis 
that, while both models use AU14 to an extent, \modelFive{} has a better representation for AU14 than \modelZero{}.
To test this hypothesis we trained pairs of SVMs to detect AU14 on multiple folds of the CK+ dataset using \modelFive{} and \modelZero{} as feature extractors (see \ref{sec:concept_embedding}).
We compared the resulting F1 scores with a paired t-test.
The SVMs trained on \modelFive{} features significantly outperformed the SVMs trained on \modelZero{} features (\modelZero{}: mean F1 $72.41\%$, \modelFive{}: mean F1 $74.13\%$, t-statistic: $-13.13$, p-value: $4.88{e}{-34}$), showing that \modelZero{} indeed forgot AU14 to a certain degree.

Regarding surprise, we observed that \modelFive{} has a stronger focus on AU5 (upper lid raise) than \modelZero{} (see figure \ref{fig:saleincy_difference_contempt} second row for example).
After inspecting the CK+ images which contain AU5, it seems like most of them also contain a  wide open mouth which corresponds to AU26 (jaw drop).
To make sure that the SVMs are trained to recognize AU5 and not AU26, we crop the bottom of the images such that they do not contain the mouth area. 
As mentioned by \cite{kim2018interpretability}, only taking cropped pictures of a concept does not hinder concept detection. 
Our evaluation shows that the SVMs trained on \modelFive{} features significantly outperformed the ones trained on \modelZero{} features (\modelZero{}: mean F1 $83.87\%$, \modelFive{}: mean F1 $84.36\%$, t-statistic: $-9.33$, p-value: $3.49\mathrm{e}{-19}$),  confirming our hypothesis that \modelZero{} forgot AU5 to a degree.

\section{Discussion}

The goal of our analysis is to find which embedded concepts are forgotten during transfer learning from emotions to pain. 
As a demonstration of our methodology, we show that \modelZero{} (fine-tuned \emph{all} convolution layers on pain dataset) pay less attention to AU5 (upper lid raise) and AU14 (dimple) in comparison to \modelFive{} (\emph{no} fine-tuning of convolution layers). 
\begin{figure}[t]
\sidecaption[t]
  \centering
  \includegraphics[width=0.649\linewidth]{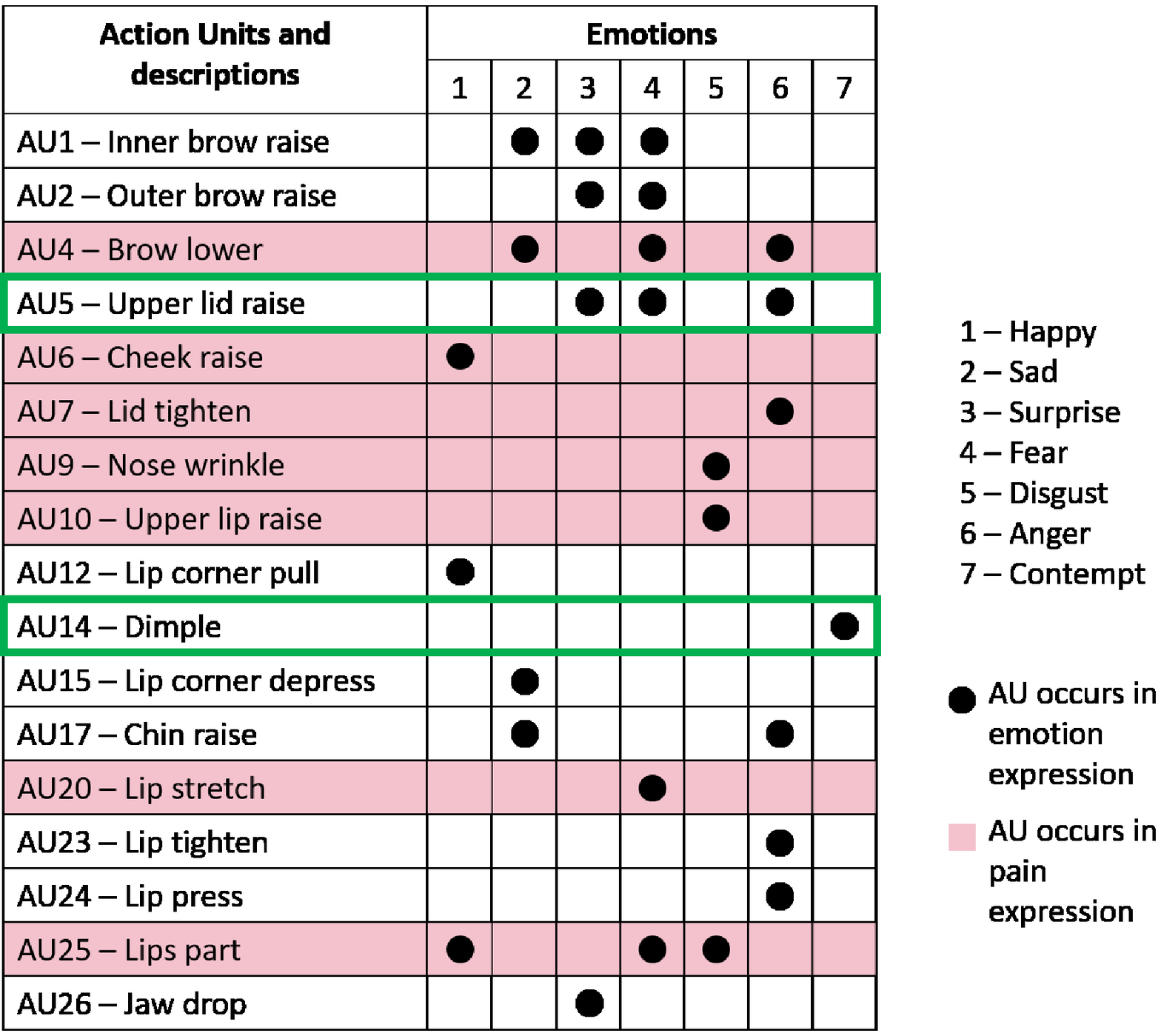}
  \caption{List of AUs that may occur in the expression of emotions and pain. The green highlighting shows the AUs that were focused less by \modelZero{} compared to \modelFive{}}
  \label{fig:AUmap}
\end{figure}
Since the forgotten embedded concepts are AUs, we investigate the relation between emotions and pain in terms of AUs. 
Figure~\ref{fig:AUmap} lists the typical AUs activated while expressing various emotions (derived from \cite{langner2010presentation}). 
The figure also highlights the AUs activated for expressing pain (according to the typical and observed activation of AUs in \cite{simon2008recognition}).
When looking at the identified AUs (AU5, AU14) that the network has forgotten during the fine-tuning process, we can see that those are used for the detection of the emotions surprise and contempt but not for pain. 
This shows that the forgetting of the semantic embedded concepts, that we identified as reason for the observed drop of recall in those two emotions (figure \ref{fig:retrainemorecall}), is in line with current theoretical findings found in literature. 

\begin{figure}[t]
\sidecaption[t]
  \centering
  \includegraphics[width=0.649\linewidth]{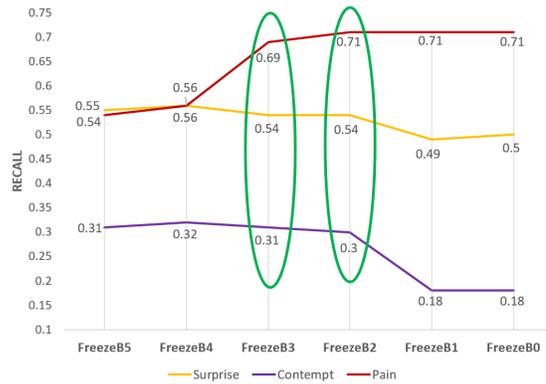}
  \caption{Plot of pain, surprise, and contempt recalls of the 6 pain models. The green ellipses highlight the best models that have a good balance between pain and emotion recalls. }
  \label{fig:recalltradeoff}
\end{figure}

Using our approach to investigate what was forgotten during transfer learning, we can conclude 
that pain training mainly affects the recognition of surprise and contempt. 
So, the trade-off between pain recognition and emotion recognition can be narrowed down to a trade-off between pain, surprise and contempt. 
Figure \ref{fig:recalltradeoff} shows the recall of pain, surprise and contempt in each of the 6 pain models (\modelZero{} to \modelFive{}). 
Pain recall is considerably lower in \modelFive{} and \modelFour{}. This is expected as these cases have lesser convolution blocks available for pain training. 
\modelOne{} and \modelZero{} yield the worst results for surprise and contempt. 
Hence, as highlighted by green ellipses in figure \ref{fig:recalltradeoff}, the best transfer learned pain recognition models that also minimises forgetting of previously learned emotion recognition are \modelThree{} and \modelTwo{}.

\section{Conclusion}
Within this paper we presented an XAI assisted approach to examine the process of knowledge transfer between two models with a focus on forgetting of already learned knowledge.
A key benefit of our approach is the framing of forgotten knowledge as semantically meaningful concepts that can be understood by humans. 
In return this enables a human observer to better understand the trade-off between an automatically learned multipurpose latent representation of data and multiple specialized representations. 
We demonstrated this process on the example of automatic pain recognition from facial expressions, using a source model that has been trained on the task of emotion recognition. 
Our results show that during the transfer learning process our model tends to pay less attention to specific action units that are known to be important markers for the emotions \textit{contempt} and \textit{surprise} but are not relevant for the problem of pain recognition.  
In the light of real world applications we argue that this knowledge about the model can be helpful to end users as well as machine learning engineers.
For the described use case of automatic pain and emotion recognition we can see an application in hospitals and nursing homes as an assistant technology for the caregivers. 
Here it is important for the staff to know about the strength and weaknesses of a model to establish trust in the technology and drive the user acceptance of the technology. 
For machine learning engineers and research scientists our process can help to understand the trade offs of an automatically learned representation. 
It also provides clues to a developer on how to improve the model further.
Given the use case of automatic pain recognition for example, it seems feasible to remove images from the \textit{contempt} and \textit{surprise} category from pre-training to potentially improve upon the results of the final model - an idea we would like to further investigate in the future.

\section{ Acknowledgments}
This work has received funding from the European Union under grant agreement number 847926 MindBot and from the DFG under project number 392401413, DEEP.

\bibliographystyle{styles/spbasic}
\bibliography{bibliography}
\end{document}